\documentclass[conference]{IEEEtran}
\IEEEoverridecommandlockouts
\usepackage{cite}
\usepackage{amsmath,amssymb,amsfonts}
\usepackage{algorithmic}
\usepackage{graphicx}
\usepackage{textcomp}
\usepackage{xcolor}
\usepackage[colorlinks,urlcolor=blue]{hyperref}
\def\BibTeX{{\rm B\kern-.05em{\sc i\kern-.025em b}\kern-.08em
    T\kern-.1667em\lower.7ex\hbox{E}\kern-.125emX}}
\usepackage{tabularx}
\usepackage{multirow}
\usepackage{bm}
\usepackage{amssymb}
\usepackage{pifont}

\begin{document}
\title{Contraband Materials Detection Within Volumetric 3D Computed Tomography Baggage Security Screening Imagery}

\author{\IEEEauthorblockN{Qian Wang}
\IEEEauthorblockA{\textit{Department of Computer Science} \\
\textit{Durham University}\\
Durham, UK \\
}
\and
\IEEEauthorblockN{Toby P. Breckon}
\IEEEauthorblockA{\textit{Department of \{Computer Science $|$ Engineering\}} \\
\textit{Durham University}\\
Durham, UK \\
}
}

\maketitle
\begin{abstract}
Automatic prohibited object detection within 2D/3D X-ray Computed Tomography (CT) has been studied in literature to enhance the aviation security screening at checkpoints. Deep Convolutional Neural Networks (CNN) have demonstrated superior performance in 2D X-ray imagery. However, there exists very limited proof how deep neural networks perform in materials detection within volumetric 3D CT baggage screening imagery. We attempt to close this gap by applying Deep Neural Networks in 3D contraband substance detection based on their material signatures. Specifically, we formulate it as a 3D semantic segmentation problem to identify material types for all voxels based on which contraband materials can be detected. To this end, we firstly investigate 3D CNN based semantic segmentation algorithms such as 3D U-Net and its variants. In contrast to the original dense representation form of volumetric 3D CT data, we propose to convert the CT volumes into sparse point clouds which allows the use of point cloud processing approaches such as PointNet++ towards more efficient processing. Experimental results on a publicly available dataset (NEU ATR) demonstrate the effectiveness of both 3D U-Net and PointNet++ in materials detection in 3D CT imagery for baggage security screening.
\end{abstract}


\section{Introduction}
Effective and efficient baggage screening at checkpoints in airports is crucial for aviation security. In most airports, X-ray machines are deployed to scan hand baggage for prohibited objects and contraband/threat materials. The reconstructed X-ray imagery is shown to human operators who will use their expertise and experience to find out potential prohibited items within the images. The task is challenging when the baggage is packed with clutter such as large electronics since the resulting inter-occlusion between objects may lead to difficulty in identifying potential threat and contraband items. For this reason, passengers are usually required to divest any large electronic devices (e.g., laptop, tablet) and liquids before security screening.
To improve the detection rate without affecting the checkpoint throughput, airports are currently increasing the use of 3D CT screening which does not require the removal of electronic devices and liquids during baggage screening. The reconstructed 3D CT images provide more information and make it possible for the human operators to inspect the 3D CT images from differing views. 

In recent years, impressive progress in deep learning techniques has enabled the possibility of fully automatic prohibited object detection in 2D X-ray imagery with high precision and very low false alarm rates \cite{bhowmik2019good,gaus2019evaluating}.
With the success of automatic threat object detection in 2D X-ray imagery, attempts have been made to extend this idea to 3D CT imagery with promising results achieved in prior work \cite{wang2020evaluation,wang2020multi}.
However, the techniques used in \cite{wang2020evaluation} and \cite{wang2020multi} rely on the detection of specific object appearance and shape (e.g., handguns, bottles, knives, etc.) hence is likely to fail in detecting contraband materials (e.g., explosive material, drugs, etc.) which can appear in arbitrary shapes. Existing research in contraband/threat material classification and detection are mainly based on traditional approaches such as morphological operations based segmentation followed by a classifier \cite{wang2019approach}. It is unknown how deep learning techniques perform in materials detection within 3D CT imagery.

To address this issue, we attempt to address the contraband material detection problem within volumetric 3D CT baggage security screening imagery. Specifically, we formulate it as a semantic segmentation problem and generate voxel-wise semantic labelling maps based on the materials. Post-processing is subsequently applied to the segmentation results to estimate the potential contraband material signatures. We use semantic segmentation methods such as the popular U-Net \cite{ronneberger2015u} architecture and its variants. Alternatively, we investigate the possibility of converting the dense volumetric 3D data to sparse point clouds and use point cloud processing methods (e.g., PointNet++ \cite{qi2017pointnet2}) for semantic segmentation. 

To evaluate the effectiveness of the proposed approach to contraband material detection in 3D CT volumes, we conduct experiments on the public Northeastern University Automatic Threat Recognition (NEU ATR) dataset \cite{atrAlert}. Experimental results demonstrate our proposed approach using 3D Convolutional Neural Networks (CNN) based U-Net architectures outperforms traditional 3D segmentation based methods \cite{wang2019approach} whilst point cloud based methods are more computationally efficient.

To summarize, the contributions of this paper are as follows:
\begin{itemize}
    \item the first attempt to address contraband materials detection within volumetric 3D CT baggage security screening imagery using various deep learning models such as 3D U-Net and PointNet++.
    \item the first attempt to convert volumetric 3D data to point clouds to promote computationally efficient processing and to compare 3D U-Net architectures and PointNet++ as two differing representation paradigms for volumetric 3D imagery processing.
    \item a framework for contraband materials detection is proposed by formulating it as a semantic segmentation problem followed by post-processing operations and is validated through extensive experiments on a publicly available dataset with three types of target materials for detection and classification (i.e. saline, rubber and clay).
\end{itemize}
\section{Related Work}\label{sec:related}
In this section, we review existing work related to our own from the perspective of \textit{baggage security screening} and \textit{3D semantic segmentation}.

\subsection{Baggage Security Screening}

Automatic object detection and recognition algorithms have been proposed and evaluated for baggage aviation security screening based on 2D X-ray images \cite{akcay2018using,bhowmik2019good}. The use of CNN architectures and object detection frameworks boosts the performance with a high detection rate and a low false-positive rate. For instance, Gaus et al. \cite{gaus2019evaluation} evaluate the effectiveness of Faster R-CNN \cite{ren2015fasterrcnn}, Mask R-CNN \cite{he2017maskrcnn} and RetinaNet \cite{lin2017focal} in detecting six different objects (i.e. bottle, hairdryer, iron, toaster mobile and laptop) in 2D X-ray baggage images.

To enable automatic baggage screening using 3D CT imagery, a variety of studies have been carried out in recent years \cite{wiley2012automatic,flitton20123d,mouton20143d,jin2015joint,flitton2015object,mouton2015materials,mouton2015review,wang2019approach,wang2020reference}.

One research direction is object segmentation based on the material and morphological structure \cite{wiley2012automatic,mouton2015materials,wang2019approach}. Specifically, Mouton et al. \cite{mouton2015materials} propose a two-stage approach for object segmentation within 3D CT imagery. A CT volume is firstly coarsely segmented based on the voxel intensity ranges of pre-defined materials. Subsequently, a variety of shape descriptors are computed as features for the random forest classifier to determine a segment resulted from the first stage is good (containing only one object) or bad (containing multiple objects and hence need further segmentation). Wang et al. \cite{wang2019approach} along with others \cite{atrAlert,paglieroni2018consensus} studied the issue of object segmentation and classification in 3D CT imagery and focused mainly on the material characteristics without considering any specific prohibited item (e.g., firearm, knife, etc.). An approach to 3D segmentation is proposed based on recursive morphological operations and the Support Vector Machines (SVM) were employed for the classification of three types of materials \cite{wang2019approach}.

3D object detection within 3D CT baggage security screening imagery has been studied in \cite{megherbi2010classifier, flitton2013comparison, wang2020evaluation}. Flitton et al. \cite{flitton2013comparison} evaluate the effectiveness of different 3D descriptors in a search-based detection approach. Their approach is limited to detect known objects for which the reference data are assumed to be available. Such an assumption hinders its application in practice when the reference data are usually unavailable. Wang et al. \cite{wang2020evaluation} use contemporary object detection frameworks based on 3D CNN and evaluate its performance on individual object detection independently.  However, most of these existing works focus on the prohibited objects of specific appearances and shapes and may not perform well for contraband items of specific materials. To close this gap, we attempt to investigate the possibility of using 3D deep learning techniques \cite{brock2016generative,zhou2018voxelnet} for contraband materials detection within volumetric 3D CT imagery for security screening.
In parallel to the research mentioned above, there also exist studies on Explosive Device Systems (EDS) for aviation security screening \cite{edic2011integrated,wells2012review,chan2017towards,jumanazarov2020system} which focus on the detection of explosive materials. 

\subsection{3D Segmentation}
3D segmentation is a typical approach to 3D scene understanding based on RGB-D data \cite{gupta2014learning,wu20153d,li2016lstm,song2017semantic}.
The first category of approaches to RGB-D semantic segmentation encoded the depth map as an image which can be processed by 2D CNN in a similar way to RGB image processing \cite{gupta2014learning,li2016lstm}. Alternatively, 3D CNN models were employed in \cite{wu20153d,song2017semantic}. The depth maps were used to represent the 2D RGB images into the corresponding 3D volumetric representations. By comparison, the CT data are naturally in the form of 3D volumes. However, 3D CNN models suffer from dealing with high-resolution data due to the computation cost. To alleviate this issue, 3D graph neural networks were proposed for RGBD semantic segmentation \cite{qi20173d}.

Medical image analysis is one of the most active areas of 3D segmentation. U-Net \cite{ronneberger2015u} is an effective architecture of deep neural network model for semantic segmentation. It has been extended to its 3D variant for a variety of applications such as pulmonary nodule segmentation \cite{funke20203d,chen2019med3d}, skin lesion segmentation \cite{ibtehaz2020multiresunet}, kidney tumor segmentation \cite{isensee2019attempt} and infant brain segmentation \cite{qamar2020variant} in medical images of varying modalities (e.g., Electron Microscopic, CT and MRI).
Specifically, MultiResUNet \cite{ibtehaz2020multiresunet} modifies the original U-Net by employing Inception-style blocks to capture multi-scale features in the contraction path (i.e. the encoder). A similar idea of Inception modules was also employed by \cite{qamar2020variant} for infant brain MRI segmentation.
Residual U-Net introduces residual modules (i.e. skip connections \cite{he2016deep}) to the CNN blocks in the U-Net architecture and was reported to outperform U-Net by \cite{isensee2019attempt}.
A transfer learning framework was proposed in \cite{chen2019med3d} to alleviate the training data sparsity issues of most medical image analysis tasks.
\begin{figure*}
    \centering
    {\includegraphics[width=0.9\textwidth]{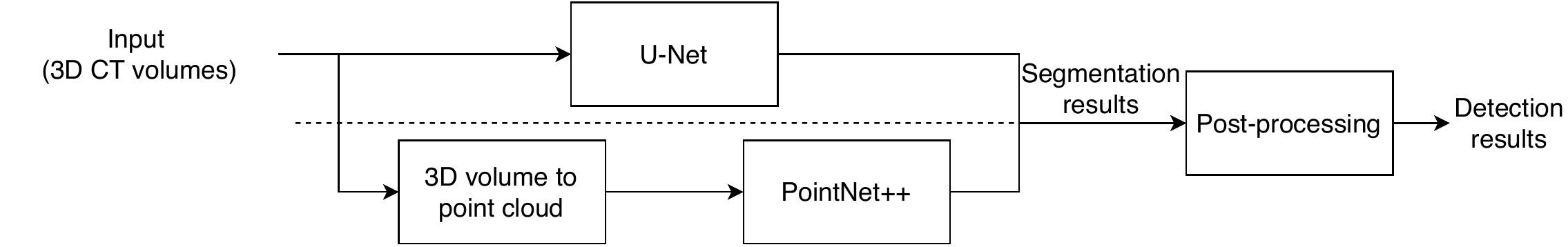}}
    {\caption{The pipeline of our approach to contraband materials detection in baggage CT volumes (two options for semantic segmentation are represented as two branches, see details in Section \ref{sec:3dcnn_method} and \ref{sec:pc_method}).}
        \label{fig:pipeline}}
\end{figure*}

Other than the applications in medical image analysis, 3D semantic segmentation has also been extensively studied in the domain of point cloud data analysis (e.g., Lidar point cloud) for a variety of applications including autonomous driving. PointNet \cite{qi2017pointnet} and its variant PointNet++ \cite{qi2017pointnet2} are two leading contemporary methods using end-to-end frameworks directly encoding the point cloud into context-aware features for different down-stream tasks including 3D semantic segmentation. Due to their generality and effectiveness demonstrated in the literature, we employ both PointNet and PointNet++ to investigate the possibility of converting 3D CT volumes to point clouds for efficient semantic segmentation. Other alternative approaches to point cloud segmentation use the idea of voxelization to transform the sparse point clouds to grid representations which can be fed into 3D CNN models \cite{hu2020randla}. This is an inverse process of our method and has the limitation of high computation cost. Since the raw data we are concerned with is in the form of 3D volumes, the other alternative to reducing the computation cost is to use sparse convolutional networks \cite{graham20183d} which is also employed by \cite{hu2020randla}. In our work, we focus mainly on the performance of different semantic segmentation methods and hence we use a simple down-sampling strategy to reduce image resolution for more efficient computation.

\section{Method} \label{sec:method}
Our work focuses on the detection of contraband materials with no specific appearances/shapes. Existing prohibited object detection works focus on the detection of prohibited objects such as firearms and knives employ traditional object detection frameworks (i.e. predicting a bounding box for each detected object) \cite{wang2020evaluation}. However, this is not an optimal choice for the detection of contraband materials which can be of arbitrary shapes (e.g., curved sheets, liquid in different containers). 
As a result, we formulate it as a {\it semantic segmentation} problem and attempt to predict voxel-wise labels for 3D baggage CT volumes. The segmentation results give potential locations of contraband items and the classes they belong to. Post-processing is subsequently employed to refine the segmentation results and generate contraband materials detection results.

In the following subsections, we first introduce two different approaches to 3D semantic segmentation. As we formulate it as a semantic segmentation problem, we first extend the prevalent U-Net architecture \cite{ronneberger2015u,cciccek20163d}  for 2D image segmentation to our 3D CT segmentation scenario. To improve efficiency and reduce the memory and processing time consumption, we also explore the possibility of using point cloud processing methods \cite{qi2017pointnet,qi2017pointnet2} for CT segmentation. To this end, we convert the volumetric 3D CT data into sparse point clouds which reserve only a small fraction of useful voxels in the original volumes as points. As illustrated in Figure \ref{fig:pipeline}, segmentation results are finally post-processed to generate contraband materials detection results.

\subsection{3D CNN Based Method}\label{sec:3dcnn_method}
\begin{figure*}
    \centering
    {\includegraphics[width=\textwidth]{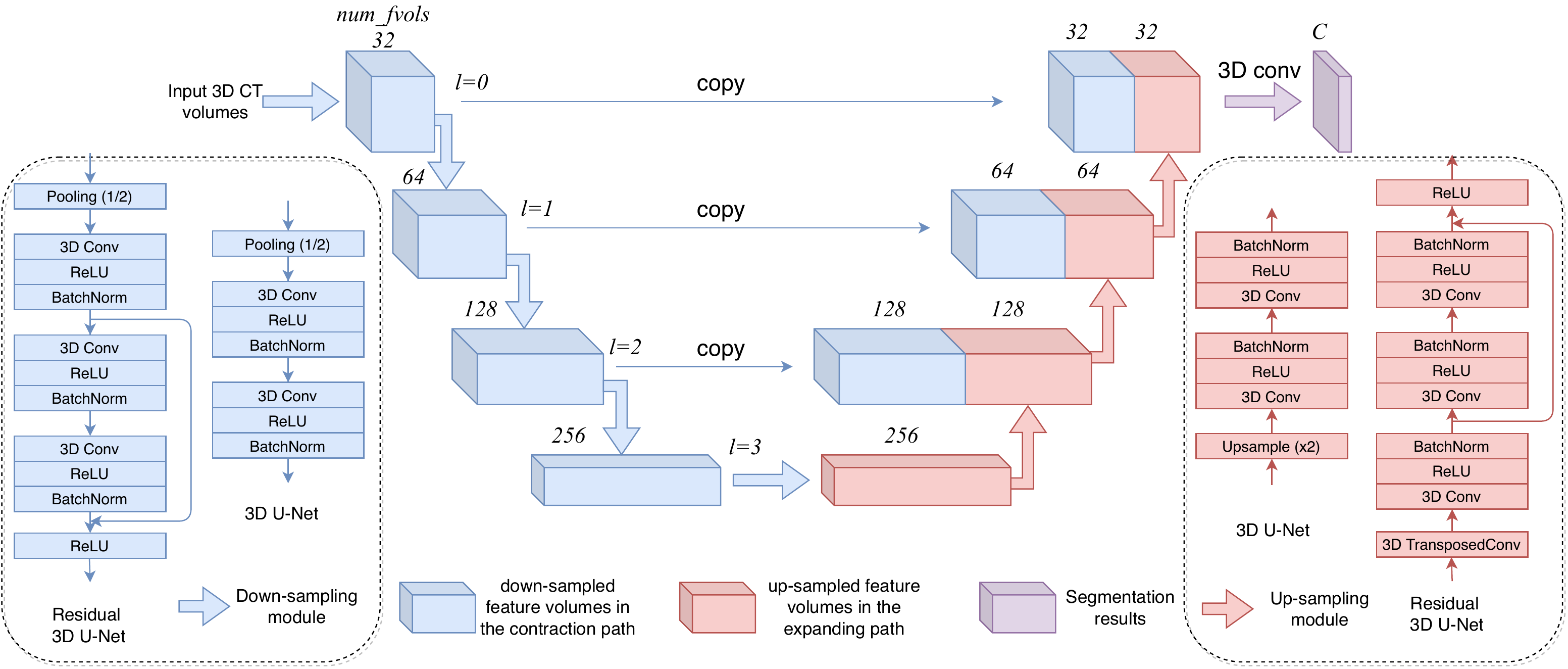}}
    {\caption{3D U-Net architectures for semantic segmentation. }
        \label{fig:3dunet}}
\end{figure*}

We follow the work in \cite{cciccek20163d} and extend U-Net architecture to 3D scenarios. As shown in Figure \ref{fig:3dunet}, 3D U-Net also consists of a contraction path and an expanding path. The contraction path is composed of a sequence of down-sampling modules to capture context from the input 3D volumes whilst the expanding path uses a sequence of upsampling modules to expand the low-resolution feature volumes extracted by the contraction path to the original resolution. 
There are $L$ down-sampling modules in the contraction path and the same number of up-sampling modules in the expanding path. Skip connections are used to copy the feature volumes from the contraction path and concatenate them with the feature volumes of the same level in the expanding path.

The down-sampling module in 3D U-Net architecture is composed of two 3D convolution layers, each of which is followed by ReLU and batch normalisation layers. A pooling layer is used to down-sampling the spatial resolution of feature volumes by a factor of 2 in all three dimensions. Specifically, the number of feature volumes in the $l$-th level is $2^l$ times more than that in the $0$-th level (denoted as $fvols$) whilst the dimension of the feature volumes is $1/2^l$ of that in the original CT volume. 
On the other hand, the up-sampling module is implemented by a non-parametric upsample function in 3D U-Net followed by two consecutive 3D convolution layers (each is followed by ReLU and batch normalisation layers) similar to the counterparts in the contraction path.
The details of down-sampling and up-sampling modules are shown in the dashed boxes of Figure \ref{fig:3dunet}.

Figure \ref{fig:3dunet} also illustrates the details of down-sampling and up-sampling modules of the residual 3D U-Net architecture in the dashed boxes. Compared with those in 3D U-Net, the main difference is three-fold. Firstly, there are three 3D convolution layers (as well as the ReLU and batch normalisation layers) in each module. Secondly, there is a skip connection between the first and third convolution layers. Finally, a 3D deconvolution layer is used in residual 3D U-Net as opposed to the interpolation-based up-sampling layer in the 3D U-Net.



\subsection{Point Cloud Based Methods}\label{sec:pc_method}
In a baggage CT volume, there are usually large regions of non-threat voxels which can be easily recognized by simple thresholding based on prior knowledge of the density ranges of these benign materials in Hounsfield units. To alleviate the memory and time-intensive issues of 3D CNN models, we attempt to investigate the possibility of point cloud processing algorithms in semantic segmentation for volumetric 3D CT data.

In the first stage, CT volumetic data is converted into a point cloud by reserving only the voxel-of-interest. Specifically, we use prior knowledge to set thresholds and consider only voxels whose intensity values are within a specified range where contraband materials fall into. As a result, a point is represented by a 4-dimensional vector of coordinates $x, y, z$ and the intensity $i$.

We use the most popular point cloud processing models PointNet \cite{qi2017pointnet} and its extension PointNet++ \cite{qi2017pointnet2} for the proof of concept as they perform the best among a few candidates in our preliminary experiments. PointNet and PointNet++ take point clouds as the input and generate segmentation results for our purpose. We use the default model architecture proposed in the original paper.

Specifically, PointNet takes $n$ points from a point cloud (may be a sub-block within a large point cloud) as input which is represented as a $n\times 4$ matrix. PointNet first transforms the input by a learnable $3\times3$ transformation matrix. Subsequently, all point features are fed into a Multi-Layer Perceptron module (MLP) and transformed to a $n\times 64$ feature matrix. Similar processing is repeated and finally a $n\times 1024$ feature matrix is generated and max-pooled to get a global feature of 1024 dimensions. For semantic segmentation in our case, the global feature is concatenated with each point feature (i.e. the one of 64 dimensions) to form a $n \time 1088$ feature matrix which again is fed into a sequence of MLP until the final output layer generating the segmentation results.

PointNet++, as an extension of PointNet, takes $n$ points from a point cloud as input which is represented as a $n\times 7$ matrix where the three more features than those used in PointNet are normalized point coordinates $x'$, $y'$ and $z'$. PointNet++ learns hierarchical point set features via the set abstraction module which is composed of point sampling and grouping followed by PointNet based feature extraction. The set abstraction modules are repeated for twice before a sequence of interpolation and unit PointNet to generate final segmentation results of the same resolution as the input.
\subsection{Post-processing} \label{sec:postproc}
The segmentation results of 3D U-Net and PointNet++ are voxel-wise and point-wise class labelling respectively. We convert these segmentation results to detection results in the post-processing stage. Specifically, we group the connected voxels which are labelled as the same class as a detected object. To these ends, we use morphological operations to correct the mislabelling information in the segmentation results.

\begin{figure}
    \centering
    {\includegraphics[width=0.3\textwidth]{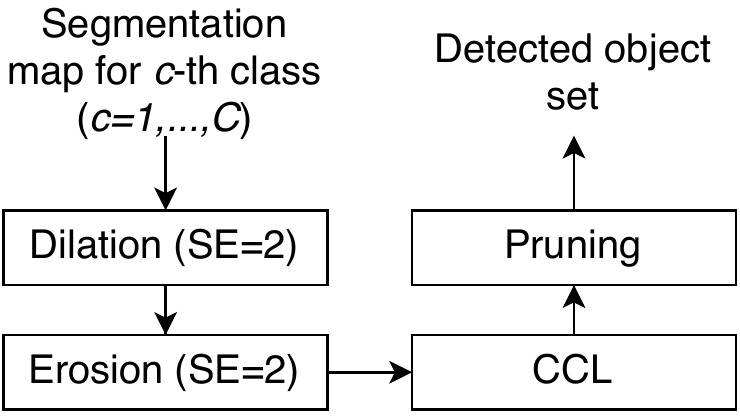}}
    {\caption{The pipeline of post-processing using morphological operations to convert segmentation results to detection results.}
        \label{fig:postproc}}
\end{figure}

The pipeline of post-processing is shown in Figure \ref{fig:postproc}. For each foreground class, we apply {\it dilation} and {\it erosion} operations sequentially to the binary segmentation map to correct the missing voxel labels within the detected objects. Subsequently, the connected component labelling (CCL) algorithm is employed to group the labelled voxels into a set of potential detected objects. We prune the detection results by removing the objects whose volumes are smaller than a pre-defined threshold.

\section{Experiments and Results}\label{sec:experiments}
In this section, we conduct experiments on a public baggage CT dataset to evaluate the effectiveness of our proposed approaches to contraband materials detection. We describe the details of the dataset and evaluation metrics used in our experiments. Subsequently, we report the quantitative results of 3D CNN methods and point cloud methods respectively. Finally, qualitative evaluation is presented to provide some intuitive insight into how our approaches perform.

\subsection{Dataset}
We follow \cite{wang2019approach} and use the Northestern University Automated Threat Recognition (NEU ATR) dataset \cite{atrAlert,atrdata} collected and annotated by NEU ALERT throughout our experiments in this study.
Baggage CT volumes were collected by a medical CT scanner (Imatron C-300). The slice size is 512$\times$512 corresponding to the field view of 475 mm$\times$475 mm hence the in-plane pixel size is 0.928 mm. The number of slices varies in different volumes and the slice spacing is 1.5 mm. Pixel values are represented by the Modified Hounsfield Unit (MHU) ranging from 0 to 32,767 MHU in which air and water are 0 and 1024 respectively.

The ATR dataset consists of 188 CT volumes in which there are 446 object signatures of three target materials (i.e. \textit{saline}, \textit{rubber} and \textit{clay}) and other non-target materials as cluttered background of typical packed baggage.  The ground truth voxels are labelled by NEU ALERT for all the objects of three target materials. We follow \cite{wang2019approach} to split the whole dataset into two subsets evenly: \textit{odd} set and \textit{even} set containing 94 odd and even indexed volumes respectively (i.e. 50/50, training/testing data split). In our experiments, we use one subset for training and the other for testing.

\subsection{Evaluation Metrics}
We use three groups of evaluation metrics in our experiments. The first one is the mean Intersection over Union (IoU) which is a typical evaluation metric for semantic segmentation as the contraband materials detection has been formulated as a semantic segmentation problem. IoU is computed for each class and the mean IoU is the mean of each IoU for all classes. The mean IoU evaluates the performance of segmentation but cannot measure the detection performance of individual objects in a CT volume.

The second group of evaluation metrics are the precision and recall which can be computed based on the detection results obtained after post-processing the segmentation results (c.f. Section \ref{sec:postproc}).  

To make a direct comparison with the traditional method proposed in \cite{wang2019approach}, we also use a third group of evaluation metrics in our experiments which have been used in \cite{wang2019approach}. These evaluation metrics are similar to typical ones for object detection (i.e. precision and recall in the second group) but concern the Probability of Detection (PD) and the Probability of False Alarms (PFA). PD is similar to recall and the main difference is that PD is computed over all detections regardless of their classes whilst recall is computed class-wisely. PD is defined in this way so that the detection model focuses more on the difference between contraband items and benign ones rather than the difference between different types of contraband items.
\begin{table}[h]
\centering
\caption{IoU results of material segmentation within 3D CT volumes on NEU ATR dataset (L -- \# of levels in 3D U-Net; fvols -- \# of feature volumes in 3D U-Net; fac. -- the downsamping factor used to down-sample the original CT volumes).}\label{table:iou}
\newsavebox{\tablebox}
\begin{lrbox}{\tablebox}
\begin{tabular}{l ccc| ccccc}
\hline
Method & L & fvols & fac. & Background & Saline & Rubber & Clay   & mIoU \\ \hline
PointNet & - & - & 2 & 99.0 & 15.7 & 15.5 & 31.0 & 40.3 \\
PointNet & - & - & 4 & 99.1 & 3.9 & 5.1 & 35.5 & 35.9 \\
PointNet++  & - & - & 2 & 98.9 & 39.8 & 28.2 & 61.9 & 57.2 \\
PointNet++  & - & - & 4
& 99.0 & 32.6 & 26.9 & 50.9 & 52.3 \\
3D U-Net & 4 & 32 & 2  & 99.5 & 65.3 & 58.8 & 74.5 & 74.5 \\
3D U-Net & 4 & 32 & 4  & 99.5 & 64.2 & 61.3 & 66.5 & 72.9 \\
3D U-Net & 4 & 32 & 8  & 99.5 & 58.7 & 48.7 & 60.7 & 66.9 \\
3D U-Net & 4 & 64 & 8  & \bf 99.6 & 61.8 & 53.3 & 64.5 & 69.8 \\
3D U-Net & 6 & 32 & 4  & \bf 99.6 & 65.5 & 62.1 & 69.2 & 74.1 \\
3D U-Net & 6 & 64 & 4  & \bf 99.6 & 64.9 & 63.0 & 72.5 & \bf 75.0 \\
Resisual 3D U-Net  & 4 & 32 & 2  & 99.5 & 56.6 & 57.7 & \bf 74.8 & 72.2 \\
Resisual 3D U-Net  & 4 & 32 & 4  & \bf 99.6 & 63.1 & 60.9 & 67.0 & 72.6 \\
Resisual 3D U-Net  & 6 & 32 & 4  & \bf 99.6 & 63.5 & 59.4 & 73.1 & 73.9 \\
Resisual 3D U-Net  & 6 & 64 & 4  & \bf 99.6 & \bf 67.4 & \bf 64.6 & 67.9 & 74.9 \\
\hline
\end{tabular}
\end{lrbox}
\scalebox{0.8}{\usebox{\tablebox}}
\end{table}

\begin{table*}[h]
\centering
\caption{Precision and recall results of material segmentation within 3D CT volumes on NEU ATR dataset (L -- \# of levels in 3D U-Net; fvols -- \# of feature volumes in 3D U-Net; fac. -- the downsamping factor used to down-sample the original CT volumes).}
\label{table:precision_recall}
\begin{lrbox}{\tablebox}
\begin{tabularx}{0.8\textwidth}{lccc|cc|cc|cc|cc}
\hline
\multirow{2}{*}{Model} & \multirow{2}{*}{L} & \multirow{2}{*}{fvols} & \multirow{2}{*}{fac.} & \multicolumn{2}{c|}{Saline}& \multicolumn{2}{c|}{Rubber}& \multicolumn{2}{c|}{Clay} &\multicolumn{2}{c}{Overall} \\ \cline{5-12}
 & & & &  P (\%) & R (\%) &  P (\%) & R (\%)&  P (\%) & R (\%)&P (\%) & R (\%) \\ \hline \hline 
PointNet & - & - & 2 & 35.0 & 62.8 & 40.4 & 56.8 & 58.2 & 73.6 & 44.5 & 64.4 \\ 
PointNet & - & - & 4 & 31.1 & 16.2 & 27.6 & 13.6 & 54.5 & 70.6 & 37.8 & 33.5 \\  
PointNet++ & - & - & 2 & 41.9 & 84.1 & 59.7 & 58.9 & 60.1 & 76.9 & 53.9 & 73.3 \\
PointNet++ & - & - & 4 & 37.8 & 73.2 & 56.8 & 51.3 & 52.5 & 79.8 & 49.0 & 68.1 \\ 
3D U-Net & 4 & 32 & 2 & 59.1 & \bf 94.6 & 82.6 & 85.8 & 76.5 & \bf 94.2 & 72.7 & \bf 91.5 \\ 
3D U-Net & 4 & 32 & 4 & 77.1 & 85.2 & 81.3 & 90.3 & 83.7 & 86.6 & 80.7 & 87.3 \\
3D U-Net & 4 & 32 & 8 & 72.4 & 80.3 & 82.2 & 80.6 & 83.8 & 80.2 & 79.5 & 80.4 \\ 
3D U-Net & 6 & 32 & 4 & 74.6 & 90.1 & 86.2 & 91.8 & 86.8 & 90.5 & 82.6 & 90.8 \\
3D U-Net & 6 & 64 & 4 & 71.8 & 91.0 & 87.4 & \bf 92.5 & \bf 94.6 & 85.8 & 84.6 & 89.8 \\
3D Residual U-Net & 4 & 32 & 2 & 67.2 & 79.1 & 82.2 & 80.6 & 79.4 & 92.6 & 76.3 & 84.1 \\
3D Residual U-Net & 4 & 32 & 4  & 78.0 & 82.8 & 83.8 & 88.1 & 93.2 & 83.0 & 85.0 & 84.7 \\
3D Residual U-Net & 6 & 32 & 4  & \bf 79.5 & 79.3 & 84.8 & 89.1 & 89.2 & 88.0 & 84.5 & 85.5 \\
3D Residual U-Net & 6 & 64 & 4  & 78.4 & 82.5 & \bf87.8 & 87.3 & 90.4 & 91.3 & \bf85.5 & 87.1 \\

\hline 
\end{tabularx}
\end{lrbox}
\scalebox{1}{\usebox{\tablebox}}
\end{table*}

\begin{table}
\centering
\caption{PD and PFA results of material segmentation within 3D CT volumes on NEU ATR dataset (L -- \# of levels in 3D U-Net; fvols -- \# of feature volumes in 3D U-Net; fac. -- the downsamping factor used to down-sample the original CT volumes).}
\label{table:pdpfa}
\begin{lrbox}{\tablebox}
\begin{tabularx}{0.62\textwidth}{lccc|cccc|c}
\hline
\multirow{2}{*}{Model} &\multirow{2}{*}{L} &\multirow{2}{*}{fvols} &\multirow{2}{*}{fac.} & \multicolumn{4}{c|}{PD (\%)} & PFA (\%) \\ \cline{5-9}
&&& &  Saline & Rubber & Clay & Overall & Overall \\ \hline \hline 
SVM \cite{wang2019approach} & - & - & - & 87 & 95 & \bf 96 & 92 & 24\\ 
PointNet & - & - & 2  & 81 & 84 & 88 & 84 & 29\\
PointNet & - & - & 4  & 38 & 41 & 80 & 50 & 13\\
PointNet++ & - & - & 2  & \bf 97 & 87 & 94 & 92 & 24 \\
PointNet++ & - & - & 4  & 92 & 80 & 86 & 86 & 22\\
3D U-Net & 4 & 32 & 2  & 95 & 94 & 92 & \bf 94 & 11 \\
3D U-Net & 4 & 32 & 4  & 86 & 96 & 86 & 90 & 6\\
3D U-Net & 4 & 32 & 8  & 81 & 84 & 81 & 82 & 5 \\ 
3D U-Net & 6 & 32 & 4  & 91 & 96 & 89 & 92 & 5 \\
3D U-Net & 6 & 64 & 4  & 91 & \bf 97 & 83 & 91 & 6 \\
3D Residual U-Net & 4 & 32 & 2  & 75 & 85 & 86 & 82 & 7 \\
3D Residual U-Net & 4 & 32 & 4  & 85 & 94 & 83 & 88 & 5 \\
3D Residual U-Net & 6 & 32 & 4  & 80 & 94 & 92 & 89 & 5 \\
3D Residual U-Net & 6 & 64 & 4  & 86 & 92 & 92 & 90 & \bf 4 \\
\hline 
\end{tabularx}
\end{lrbox}
\scalebox{0.75}{\usebox{\tablebox}}
\end{table}

\subsection{Experimental Settings} \label{sec:exp_settings}
Extensive experiments are conducted to evaluate the effectiveness of the proposed methods and to investigate how different factors affect the performance. 
\subsubsection{Model architecture}
We investigate two types of 3D CNN models (i.e. 3D U-Net and residual 3D U-Net) and one point cloud model (i.e. PointNet++) in our experiments. For 3D U-Net and residual 3D U-Net architectures, we consider varying depth and width by setting the number of feature volumes in the first level ($l$=0) as $fvols \in \{32, 64\}$ and the number of down-sampling and up-sampling modules as $L \in \{4,6\}$ based on the Pytorch implementation of \cite{cciccek20163d}. As a result, we can have a number of candidate models with varying combinations of settings. Instead of investigating all the possible combinations, we select typical ones performing better as shown in Tables \ref{table:iou}-\ref{table:pdpfa} and ignore those providing less insight.
In all experiments, we set the learning rate to $1e-4$ and decrease it every 50 epochs by a factor of 0.5. The training is terminated after 250 epochs.  During training, we randomly crop 3D sub-volumes of size $64\times 96 \times 96$ and apply data augmentation including normalisation, random flipping and random rotation (by 90 degrees). Each training batch constitutes of 16 such sub-volumes from cropped 4 baggage volumes (4 from each).

For PointNet and PointNet++, we use the default settings in the PyTorch implementation \cite{pointnet2code} except the input feature dimension is adapted since we have only one intensity feature in our CT data as opposed to three RGB values. The learning rate is set to $1e-3$ throughout our experiments and decays by a factor of 0.7 every 50 epochs. The training stops after 250 epochs. In the training, we randomly select 8092 points from a set of points within a block of size $48 \times 48 \times 48$. The key to training PointNet and PointNet++ is to balance the training samples of different classes. Since the background points are the majority in the training data, we need to give more weights to points belonging to foreground classes during selection (i.e. make it more likely to select the block containing foreground class points). To this end, we tend to select the blocks containing more than 50\% foreground class points as training samples.

\subsubsection{Data down-sampling}
We also investigate how the down-sampling factor affects the performance of our methods. Specifically, we down-sample the CT volumes uniformly by the factor of 2, 4 and 8 in all three dimensions respectively. During training, the ground truth labelling volumes are correspondingly down-sampled. In the evaluation, to make different results comparable, we up-sample the predicted low-resolution results to the original size and calculate the evaluation metrics.

\subsection{Experimental Results} \label{sec:results}
As the key component of our proposed approach to contraband materials detection, different semantic segmentation models are evaluated by calculating the per-class IoU and mean IoU. As intermediate results, the performance of segmentation determines the final detection performance to a large extent. The IoU results of different models are reported in Table \ref{table:iou}. The majority of voxels belong to the background class hence the IoU is close to 100\% for all methods. For three foreground classes, 3D U-Net architectures outperform point cloud based methods significantly with the best mean IoU of 75.0\%. The use of skip connection in residual 3D U-Net models does not make a difference from those without skip connections. Increasing the width and depth of the architectures of 3D U-Net benefit the performance consistently. For point cloud based methods, PointNet++ outperforms PointNet significantly. The best point cloud based method is PointNet++ with the down-sampling factor of 2 and achieves the mean IoU of 57.2\%.

Detection results are evaluated after post-processing by calculating the metrics of precision/recall. The results are shown in Table \ref{table:precision_recall} and Table \ref{table:pdpfa} respectively. Similar conclusions can be drawn from results in Table \ref{table:precision_recall} and Table \ref{table:iou}. The best overall precision and recall is achieved by 3D U-Net with the depth of 6 and 64 feature volumes in the first level and the overall precision and recall are 84.6\% and 89.8\% respectively. 

Following previous works in \cite{wang2019approach}, we report the results of PD and PFA in Table \ref{table:pdpfa}. By comparing with the results of \cite{wang2019approach}, both the point cloud based methods and 3D U-Net based methods can achieve comparable or better performance. Although PointNet gives worse results than the traditional method, its variant PointNet++ can achieve comparable PD of 92\% and PFA of 24\%. Consistent with previous results, 3D U-Net with the depth of 6 and 32 feature volumes in the first level gives the best performance with the PD of 92\% and a much less PFA of 5\%.  

\begin{figure}
    \centering
    {\includegraphics[width=0.5\textwidth]{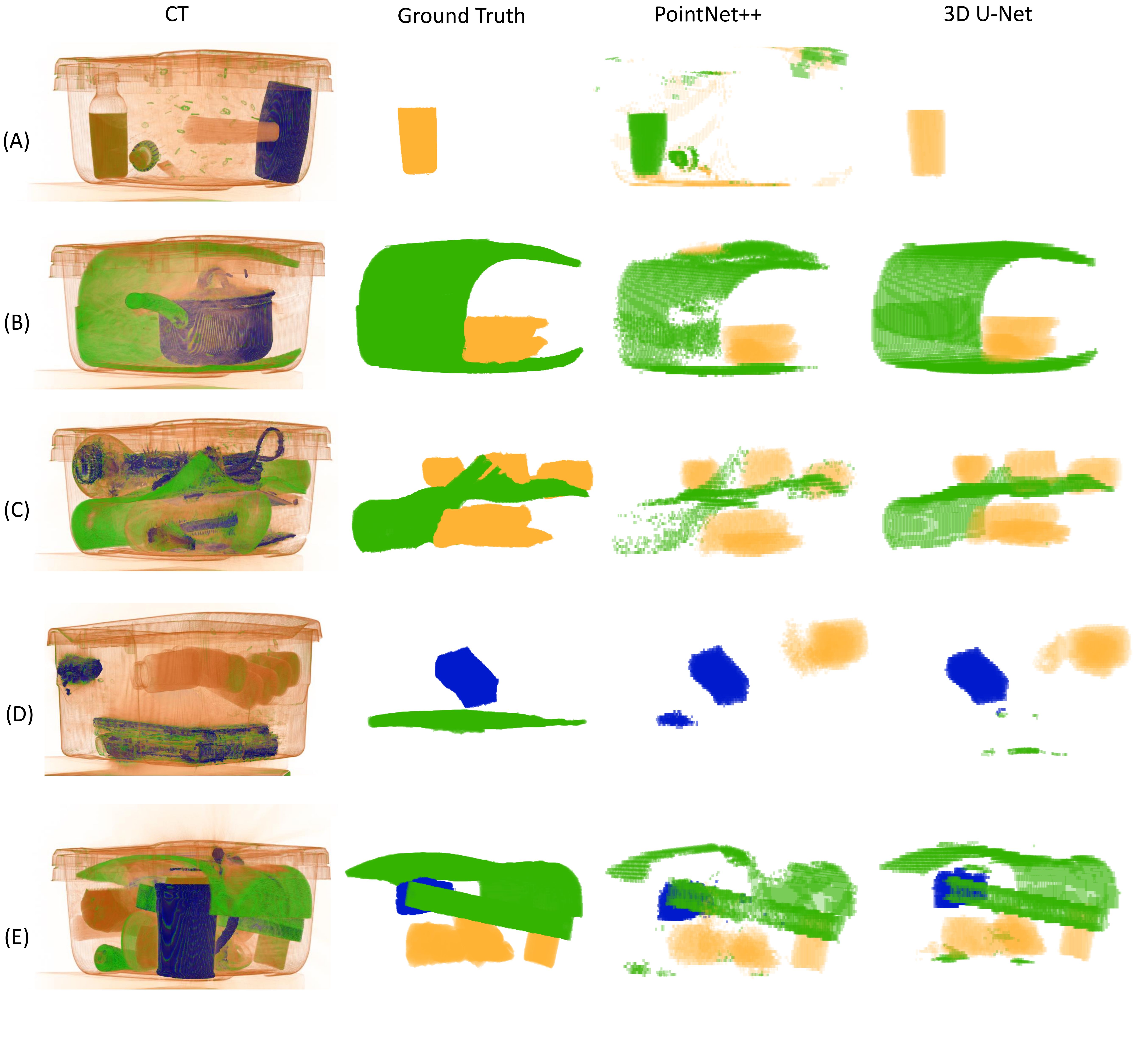}}
    {\caption{Qualitative evaluation of material segmentation and classification using varying methods for examples A-E.}
        \label{fig:visualization}}
\end{figure}

\subsection{Qualitative Evaluation} \label{sec:qualitative}
Figure \ref{fig:visualization} presents five exemplar samples of CT volumes from the ATR dataset together with their ground truth labelling and segmentation results generated by the best PointNet++ and 3D U-Net models in our experiments. The first two columns (from the left side) list the visualisations of CT volumes and their corresponding ground truth labelling. Three types of materials saline, rubber and clay are represented by \textit{orange}, \textit{green} and \textit{blue} colours respectively. The last two columns list the segmentation results of PointNet++ and 3D U-Net respectively.  It can be seen from example (A) that some background voxels are mistakenly classified as foreground classes which will lead to low precision in Table \ref{table:precision_recall} and high PFA in Table \ref{table:pdpfa}. In addition, PointNet++ mistakenly classifies the material \textit{saline} as \textit{rubber}. The examples (B) and (C) shows that PointNet++ misses a considerable amount of voxels belonging to a \textit{rubber} sheet (green) but 3D U-Net gives much better results. There also exist cases where both PointNet++ and 3D U-Net fail to detect the rubber sheet in example (D) but mistakenly detect a false alarm \textit{saline} object.

\subsection{Computational Complexity}\label{sec:complexity}
Table \ref{table:complexity} presents the computational complexity of different models investigated in our work. We consider the number of parameters and floating point operations (FLOP) given a typical 3D CT volume with a size of $300\times 512 \times 512$ in NEU ATR dataset. From Table \ref{table:complexity}, we can draw the following conclusions. Firstly, PointNet++ not only performs better but also has fewer parameters and computational cost than PointNet. Secondly, 3D U-Net is more efficient than 3D residual U-Net since it has fewer parameters but comparable performance when the depth and width are the same. Finally, increasing the depth and width of 3D U-Net is not efficient given the marginal performance gain and significantly increased computational cost. All these conclusions provide insightful instructions for our future work on volumetric 3D CT segmentation.

\begin{table}
\centering
\caption{Computational complexity of different models.}
\label{table:complexity}
\begin{lrbox}{\tablebox}
\begin{tabularx}{0.5\textwidth}{lccc|rr}
\hline
Model & L & fvols & fac. & \# Parameters (M) & FLOP (G) \\  \hline \hline 
PointNet & - & - & 2  & 3.53 & ~3440 \\
PointNet & - & - & 4  & 3.53 & ~430 \\
PointNet++ & - & - & 2  & \bf 0.97 & ~460 \\
PointNet++ & - & - & 4  & \bf 0.97 & ~57\\
3D U-Net & 4 & 32 & 2  & 4.08 & ~2250 \\
3D U-Net & 4 & 32 & 4  & 4.08 & ~280\\
3D U-Net & 4 & 32 & 8  & 4.08 & \bf ~35 \\ 
3D U-Net & 6 & 32 & 4  & 51.86 & ~305 \\
3D U-Net & 6 & 64 & 4  & 207.43 & ~1220 \\
3D Residual U-Net & 4 & 32 & 2  & 8.77 & ~3520 \\
3D Residual U-Net & 4 & 32 & 4  & 8.77 & ~440 \\
3D Residual U-Net & 6 & 32 & 4  & 84.87 & ~470 \\
3D Residual U-Net & 6 & 64 & 4  &339.44 & ~1880 \\
\hline 
\end{tabularx}
\end{lrbox}
\scalebox{1}{\usebox{\tablebox}}
\end{table}

\section{Conclusion}
In this work, we have investigated two different deep learning methods for contraband materials detection in volumetric 3D CT imagery. It is demonstrated that both 3D CNN and point cloud based methods can give reasonably good results for this task and 3D U-Net outperforms PointNet++ in terms of varying evaluation metrics. However, the point cloud based methods provide an alternative solution to the efficient processing of large 3D CT volumes and are worthy of further investigations.

In the future, we would like to expand the evaluation dataset and consider more types of contraband materials in aviation security screening. Besides, the detection performance can be enhanced by employing more advanced architectures \cite{yan2020pointasnl} based on either 3D U-Net or point cloud processing algorithms. Finally, it will be of great value to integrate the detection of the prohibited object based on appearances \cite{wang2020evaluation} and contraband items based on materials in a unified framework for plausible real-world applications.  
\ifCLASSOPTIONcaptionsoff
  \newpage
\fi

\bibliographystyle{IEEEtran}
\bibliography{ref}
\end{document}